\documentclass[11pt]{article}

\usepackage{acl2013}
\usepackage{times}
\usepackage{url}
\usepackage{latexsym}
\usepackage{paralist}
\usepackage{graphicx}
\usepackage[utf8]{inputenc}
\usepackage[russian,arabic,english]{babel}
\usepackage{multirow}
\usepackage{array}
\newcolumntype{H}{>{\lrbox0}c<{\endlrbox}@{}}
\usepackage{subfigure}
\usepackage{rotating}

\title{Polyglot: Distributed Word Representations for Multilingual NLP}
\author{Rami Al-Rfou \And
  Bryan Perozzi \\
  Computer Science Dept. Stony Brook University Stony Brook, NY 11794\\
  {\tt \{ralrfou, bperozzi, skiena\}@cs.stonybrook.edu} \\
   \And Steven Skiena
 }

\date{}

\begin{document}
\maketitle
\begin{abstract}
Distributed word representations (word embeddings) have recently contributed to competitive performance in language modeling and several NLP tasks.
In this work, we train word embeddings for more than 100 languages using their corresponding Wikipedias.
We quantitatively demonstrate the utility of our word embeddings by using them as the sole features for training a part of speech tagger for a subset of these languages.
We find their performance to be competitive with near state-of-art methods in English, Danish and Swedish.
Moreover, we investigate the semantic features captured by these embeddings through the proximity of word groupings.
We will release these embeddings publicly to help researchers in the development and enhancement of multilingual applications.
\end{abstract}

\section{Introduction}

Building multilingual processing systems is a challenging task.
Every NLP task involves different stages of preprocessing and calculating intermediate representations that will serve as features for later stages.
These stages vary in complexity and requirements for each individual language.
Despite recent momentum towards developing multilingual tools \cite{CoNLL2007,CoNLL2009,CoNLL2012}, most of NLP research still focuses on rich resource languages.
Common NLP systems and tools rely heavily on English specific features and they are infrequently tested on multiple datasets.
This makes them hard to port to new languages and tasks \cite{Blitzer06domain}.

A serious bottleneck in the current approach for developing multilingual systems is the requirement of familiarity with each language under consideration.
These systems are typically carefully tuned with hand-manufactured features designed by experts in a particular language.
This approach can yield good performance, but tends to create complicated systems which have limited portability to new languages, in addition to being hard to enhance and maintain.

Recent advancements in unsupervised feature learning present an intriguing alternative.
Instead of relying on expert knowledge, these approaches employ automatically generated task-independent features (or word embeddings) given large amounts of plain text.
Recent developments have led to state-of-art performance in several NLP tasks such as language modeling \cite{bengio2006neural,mikolov2010recurrent}, and syntactic tasks such as sequence tagging \cite{Collobert:2011:NLP}. These embeddings are generated as a result of training ``deep" architectures, and it has been shown that such representations are well suited for domain adaptation tasks \cite{large_sentiment,marginalized_sdae}.

We believe two problems have held back the research community's adoption of these methods.
The first is that learning representations of words involves huge computational costs.
The process usually involves processing billions of words over weeks.
The second is that so far, these systems have been built and tested mainly on English.

In this work we seek to remove these barriers to entry by generating word embeddings for over a hundred languages using state-of-the-art techniques.
Specifically, our contributions include:

\begin{itemize}
\item \textbf{Word embeddings} - We will release word embeddings for the hundred and seventeen languages that have more than 10,000 articles on Wikipedia. Each language's vocabulary will contain up to 100,000 words.
The embeddings will be publicly available at (\url{www.cs.stonybrook.edu/~dsl}), for the research community to study their characteristics and build systems for new languages.
We believe our embeddings represent a valuable resource because they contain a minimal amount of normalization. For example, we do not lower case words for European languages as other studies have done for English. This preserves features of the underlying language.

\item \textbf{Quantitative analysis} - We investigate the embedding's performance on a part-of-speech (PoS) tagging task, and conduct qualitative investigation of the syntactic and semantic features they capture.
Our experiments represent a valuable chance to evaluate distributed word representations for NLP as the experiments are conducted in a consistent manner and a large number of languages are covered.
As the embeddings capture interesting linguistic features, we believe the multilingual resource we are providing gives researchers a chance to create multilingual comparative experiments.

\item \textbf{Efficient implementation} - Training these models was made possible by our contributions to Theano (machine learning library \cite{theano}). These optimizations empower researchers to produce word embeddings under different settings or for different corpora than Wikipedia.
\end{itemize}

The rest of this paper is as follows.
In Section \ref{related}, we give an overview of semi-supervised learning and learning representations related work.
We then describe, in Section \ref{embeddings}, the network used to generate the word embeddings and its characteristics.
Section \ref{corpus} discusses the details of the corpus collection and preparation steps we performed.
Next, in Section \ref{training}, we discuss our experimental setup and the training progress over time.
In Section \ref{analysis} we discuss the semantic features captured by the embeddings by showing examples of the word groupings in multiple languages.
Finally, in Section \ref{results} we demonstrate the quality of our learned features by training a PoS tagger on several languages and then conclude.

\section{Related Work}
\label{related}
There is a large body of work regarding semi-supervised techniques which 
integrate unsupervised feature learning with discriminative learning methods to improve the performance of NLP applications.
Word clustering has been used to learn classes of words that have similar semantic features to improve language modeling \cite{brown1992class} and knowledge transfer across languages \cite{cross_clusters}.
Dependency parsing and other NLP tasks have been shown to benefit from such a large unannotated corpus \cite{Koo08simplesemi-supervised}, and a variety of unsupervised feature learning methods have been shown to unilaterally improve the performance of supervised learning tasks \cite{turian2010word}.
\cite{cross_embeddings} induce distributed representations for a pair of languages jointly, where a learner can be trained on annotations present in one language and applied to test data in another.

\begin{table*}[!ht]
\begin{center}
\begin{tabular}{lll l l l l}
 	Apple & apple & Bush& bush & corpora & dangerous\\ \hline
	Dell & tomato & Kennedy& jungle & notations& costly\\
	Paramount & bean& Roosevelt& lobster & digraphs& chaotic\\
	Mac & onion & Nixon& sponge& usages & bizarre\\ 
	Flex & potato & Fisher& mud & derivations& destructive

\end{tabular}
\end{center}
\caption{Words nearest neighbors as they appear in the English embeddings.}
\label{table:case}
\end{table*}

Learning distributed word representations is a way to learn effective and meaningful information about words and their usages.
They are usually generated as a side effect of training parametric language models as probabilistic neural networks.
Training these models is slow and takes a significant amount of computational resources \cite{bengio2006neural,LargeBrain}.
Several suggestions have been proposed to speed up the training procedure, either by changing the model architecture to exploit an algorithmic speedup \cite{mnih2009scalable,morin2005hierarchical} or by estimating the error by sampling \cite{bengio2008adaptive}.

\cite{collobert:2008} shows that word embeddings can almost substitute NLP common features on several tasks.
The system they built, SENNA, offers  part of speech tagging, chunking, named entity recognition, semantic role labeling and dependency parsing \cite{Collobert_deeplearning}.
The system is built on top of word embeddings and performs competitively compared to state of art systems. In addition to pure performance, the system has a faster execution speed than comparable NLP pipelines \cite{speedread}.

To speed up the embedding generation process, SENNA embeddings are generated through a procedure that is different from language modeling.
The representations are acquired through a model that distinguishes between phrases and corrupted versions of them.
In doing this, the model avoids the need to normalize the scores across the vocabulary to infer probabilities.
\cite{expressive} shows that the embeddings generated by SENNA perform well in a variety of term-based evaluation tasks.
Given the training speed and prior performance on NLP tasks in English, we generate our multilingual embeddings using a similar network architecture to the one SENNA used.

However, our work differs from SENNA in the following ways.
First, we do not limit our models to English, we train embeddings for a hundred and seventeen languages.
Next, we preserve linguistic features by avoiding excessive normalization to the text. For example, our English model places ``\emph{Apple}" closer to IT companies and ``\emph{apple}" to fruits.
More examples of linguistic features preserved by our model are shown in Table \ref{table:case}.
This gives us the chance to evaluate the embeddings performance over PoS tagging without the need for manufactured features.
Finally, we release the embeddings and the resources necessary to generate them to the community to eliminate any barriers.

Despite the progress made in creating distributed representations, combining them to produce meaning is still a challenging task.  Several approaches have been proposed to address feature compositionality for semantic problems such as paraphrase detection \cite{SocherEtAl2011:PoolRAE}, and sentiment analysis \cite{SocherEtAl2012:MVRNN} using word embeddings.

\section{Distributed Word Representation}
\label{embeddings}
Distributed word representations (word embeddings) map the index of a word in a dictionary to a feature vector in high-dimension space.
Every dimension contributes to multiple concepts, and every concept is expressed by a combination of subset of dimensions.
Such mapping is learned by back-propagating the error of a task through the model to update random initialized embeddings.
The task is usually chosen such that examples can be automatically generated from unlabeled data (i.e so it is unsupervised). 
In case of language modeling, the task is to predict the last word of a phrase that consists of $n$ words.

\begin{figure}
\begin{raggedleft}
\includegraphics[scale=0.4]{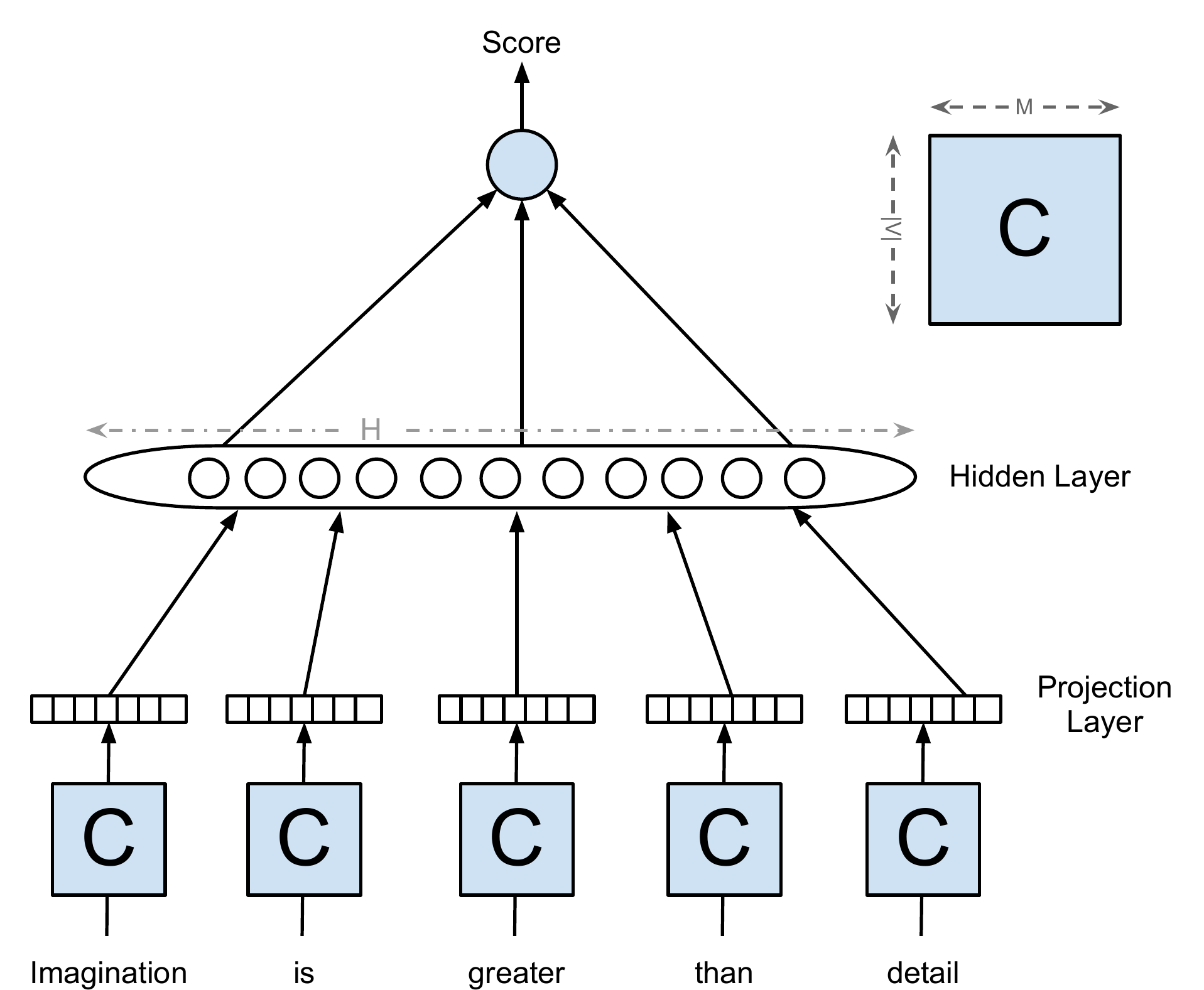}
\end{raggedleft}

\caption{Neural network architecture. Words are retrieved from embeddings matrix $C$ and concatenated at the projection layer as an input to computer the hidden layer activation. The score is the linear combination of the activation values of the hidden layer. The scores of two phrases are ranked according to hinge loss to distinguish the corrupted phrase from the original one.}
\label{model}
\end{figure}

In our work, we start from the example construction method outlined in \cite{bengio:2009}.
They train a model by requiring it to distinguish between the original phrase and a corrupted version of the phrase.
If it does not score the original one higher than the corrupted one (by a margin), the model will be penalized.
More precisely, for a given sequence of words $S = [w_{i-n} \dots w_i \dots w_{i+n}]$ observed in the corpus $T$, we will construct another corrupted sequence $S^{\prime}$ by replacing the word in the middle $w_i$ with a word $w_j$ chosen randomly from the vocabulary.
The neural network represents a function $score$ that scores each phrase, the model is penalized through the hinge loss function $J(T)$ as shown in \ref{loss}.
\begin{equation}
J(T) = \frac{1}{|T|} \sum_{i \in T} | 1- score(S^{\prime}) + score(S)|_+
\label{loss}
\end{equation}
Figure \ref{model} shows a neural network that takes a sequence of words with size $2n+1$ to compute a score.
First, each word is mapped through a vocabulary dictionary with the size $|V|$ to an index that is used to index a shared matrix $C$ with the size $\left\vert V\right\vert * M$ where $M$ is the size of the vector representing the word.
Once the vectors are retrieved, they are concatenated into one vector called projection layer $P$ with size $(2n+1) * M$.
The projection layer plays the role of an input to a hidden layer with size $|H|$, the activations $A$ of which are calculated according to equation \ref{activation}, where $W_1$, $b_1$ are the weights and bias of the hidden layer.

\begin{equation}
A = tanh(W_1 P + b_1)
\label{activation}
\end{equation}

To calculate the phrase score, a linear combination of the hidden layer activations $A$ is computed using $W_2$ and $b_2$.
\begin{equation}
score(P) = W_2 A + b_2
\label{activation}
\end{equation}

Therefore, the five parameters that have to be learned are $W_1$, $W_2$, $b_1$, $b_2$, $C$ with a total number of parameters $(2n+1) * M * H + H + H + 1 + |V|*M  \approx M*(nH + |V|)$ .

\section{Corpus Preparation}
\label{corpus}
We have chosen to generate our word embeddings from Wikipedia.
In addition to size, there are other desirable properties that we wish for the source of our language model to have:
\begin{compactitem}
\item \textbf{Size and variety of languages} - As of this writing (April, 2013), 42 languages had more than 100,000 article pages, and 117 languages had more than 10,000 article pages.
\item \textbf{Well studied} - Wikipedia is a prolific resource in the literature, and has been used for a variety of problems. Particularly, Wikipedia is well suited for multilingual applications \cite{babelnet}.
\item \textbf{Quality} - Wikipedians strive to write articles that are readable, accurate, and consist of good grammar.
\item \textbf{Openly accessible} - Wikipedia is a resource available for free use by researchers
\item \textbf{Growing} - As technology becomes more accessible, the size and scope of the multilingual Wikipedia effort continues to expand.
\end{compactitem}

To process Wikipedia markup, we first extract the text using a modified version of the Bliki engine\footnote{Java Wikipedia API (Bliki engine) - \url{http://code.google.com/p/gwtwiki/}}.
Next we must tokenize the text.
We rely on an OpenNLP probabilistic tokenizer whenever possible, and default to the Unicode text segmentation\footnote{\url{http://www.unicode.org/reports/tr29/}} algorithm offered by Lucene when we have no such OpenNLP model.
After tokenization, we normalize the tokens to reduce their sparsity.
We have two main normalization rules.
The first replaces digits with the symbol \#, so ``1999" becomes \#\#\#\#.
In the second, we remove hyphens and brackets that appear in the middle of a token.  As an additional rule for English, we map non-Latin characters to their unicode block groups.

In order to capture the syntactic and semantic features of words, we must observe each word several times in each of its valid contexts.
This requirement, when combined with the Zipfian distribution of words in natural language, implies that learning a meaningful representation of a language requires a huge amount of unstructured text.
In practice we deal with this limitation by restricting ourselves to considering the most frequently occurring tokens in each language.

\begin{table}[ht]

\begin{tabular}{|l|r|r|r|}
\hline
\multirow{2}{*}{\textbf{Language}} & \multicolumn{1}{c|}{\textbf{Tokens}} & \multicolumn{1}{c|}{\textbf{Words}} & \multirow{2}{*}{\textbf{Coverage}} \\
 & \multicolumn{1}{c|}{$* 10^6$} & \multicolumn{1}{c|}{$* 10^3$} &  \\ \hline
English & 1,888 & 12,125 & 96.30\% \\ 
German & 687 & 9,474 & 91.78\% \\ 
French & 473 & 4,675 & 95.78\% \\ 
Spanish & 399 & 3,978 & 96.07\% \\ 
Russian & 328 & 5,959 & 90.43\% \\
Italian & 322 & 3,642 & 95.52\% \\
Portuguese & 197 & 2,870 & 95.68\% \\ 
Dutch & 197 & 3,712 & 93.81\% \\ 
Chinese & 196 & 423 & 99.67\% \\
Swedish & 101 & 2,707 & 92.36\% \\
Czech & 80 & 2,081 & 91.84\% \\
Arabic & 52 & 1,834 & 91.78\% \\
Danish & 44 & 1,414 & 93.68\% \\
Bulgarian & 39 & 1,114 & 94.35\% \\
Slovene & 30 & 920 & 94.42\% \\
Hindi & 23 & 702 & 96.25\% \\ \hline
\end{tabular}

\caption{Statistics of a subset of the languages processed. The second column reports the number of tokens found in the corpus in millions while the third column reports the word types found in thousands. The coverage indicates the percentage of the corpus that will be matching words in a vocabulary consists of the most frequent 100 thousand words.}
\label{table:langs}
\end{table}

Table \ref{table:langs} shows the size of each language corpus in terms of tokens, number of word types and coverage of text achieved by building a vocabulary out of the most frequent 100,000 tokens, $|V|$. Out of vocabulary (OOV) words are replaced with a special token $\langle$UNK$\rangle$.

While Wikipedia has 284 language specific encyclopedias, only five of them have more than a million articles.
The size drops dramatically, such that the 42\textsuperscript{nd} largest Wikipedia, Hindi, has slightly above 100,000 articles and the 100\textsuperscript{th}, Tatar, has slightly over 16,000 articles\footnote{\url{http://meta.wikimedia.org/w/index.php?title=List_of_Wikipedias&oldid=5248228}}.

Significant Wikipedias in size have a word coverage over 92\% except for German, Russian, Arabic and Czech which shows the effect of heavy usage of morphological forms in these languages on the word usage distribution.

The highest word coverage we achieve is unsurprisingly for Chinese.
This is expected given the limited size vocabulary of the language - the number of entries in the Contemporary Chinese Dictionary are estimated to be 65 thousand words \cite{cn}.

\section{Training}
\label{training}

\begin{figure}[htb]
\begin{raggedleft}
\includegraphics[trim=1.0cm 0.3cm 0.0cm 0.90cm, clip=true, scale=0.45]{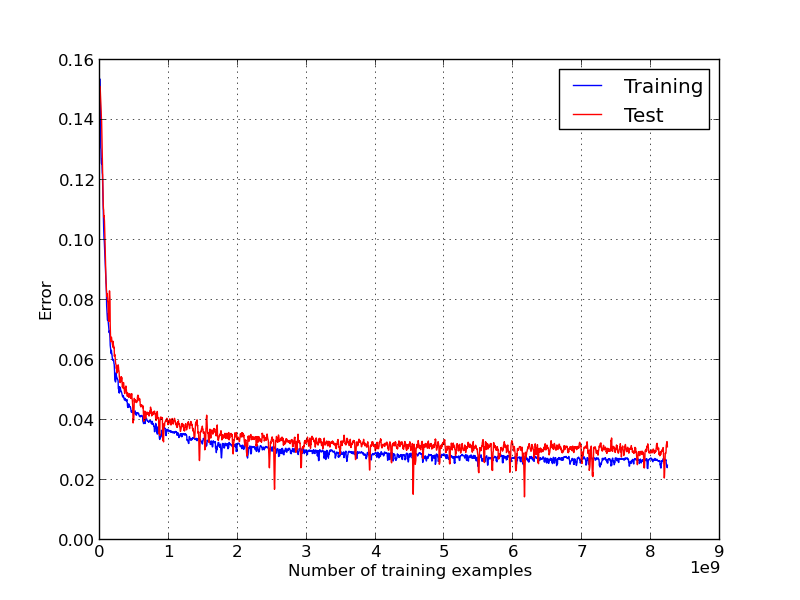}
\end{raggedleft}

\caption{Training and test errors of the French model after 23 days of training. We did not notice any overfitting while training the model. The error curves are smoother the larger the language corpus is.}
\label{error}
\end{figure}

For our experiments, we build a model as the one described in Section \ref{embeddings} using Theano \cite{theano}. We choose the following parameters, context window size $2n+1 = 5$, vocabulary $|V| = 100,000$, word embedding size $M = 64$, and hidden layer size $H = 32$. The intuition, here, is to maximize the relative size of the embeddings compared to the rest of the network.
This might force the model to store the necessary information in the embeddings matrix instead of the hidden layer.
Another benefit is that we will avoid overfitting on the smaller Wikipedias.
Increasing the window size or the embedding size slows down the training speed, making it harder to converge within a reasonable time.

The examples are generated by sweeping a window over sentences.
For each sentence in the corpus, all unknown words are replaced with a special token $\langle$UNK$\rangle$ and sentences are padded with $\langle$S$\rangle$, $\langle$/S$\rangle$ tokens.
In case the window exceeds the edges of a sentence, the missing slots are filled with our padding token, $\langle$PAD$\rangle$.

To train the model, we consider the data in mini-batches of size 16.
Every 16 examples, we estimate the gradient using stochastic gradient descent \cite{bottou-91c}, and update the parameters which contributed to the error using backpropagation \cite{rumelhart2002learning}. 
Calculating an exact gradient is prohibitive given that the dataset size is in millions of examples.
We calculate the development error by sampling randomly 10000 mini-batches from the development dataset.

For each language, we set the batch size to 16 examples, and the learning rate to be 0.1.
Following, \cite{Collobert:2011:NLP}'s advice, we divide each layer by the \emph{fan in} of that layer, and we consider the embeddings layer to have a fan in of 1.
We divide the corpus to three sets, training, development and testing with the following percentages 90, 5, 5 respectively.

\begin{table*}[!htb]
\begin{scriptsize}
\begin{center}

\begin{tabular}{lllllllll}
\cline{2-3}\cline{5-6}\cline{8-9}
& \textbf{Word} & \textbf{Translation} &   & \textbf{Word} & \textbf{Translation} & &\textbf{Word}&\textbf{Word}\\
\cline{2-3}\cline{5-6}\cline{8-9}
\multirow{7}{*}{\begin{sideways}\textbf{French}\end{sideways}}
& \textbf{rouge} & \textbf{red} &
\multirow{7}{*}{\begin{sideways}\textbf{Spanish}\end{sideways}}
& \textbf{dentista} & \textbf{dentist} &\multirow{7}{*}{\begin{sideways}\textbf{English}\end{sideways}}& \textbf{Mumbai} & \textbf{Bombay}\\

& juane & yellow &  & peluquero & barber &&Chennai&Madras\\
& rose & pink & &ginecólog & gynecologist &&Bangalore&Shanghai\\
& blanc & white & &camionero & truck driver &&Kolkata&Calultta\\
& orange & orange & & oftalmólogo & ophthalmologist&&Cairo&Bangkok\\ 
& bleu & blue & & telegrafista & telegraphist&&Hyderabad&Hyderabad\\ 
\cline{2-3}\cline{5-6}\cline{8-9}

\multirow{8}{*}{\begin{sideways}\textbf{Arabic}\end{sideways}}& \multicolumn{2}{c}{}   & \multirow{8}{*}{\begin{sideways}\textbf{Arabic}\end{sideways}}&  \multicolumn{2}{c}{} &\multirow{8}{*}{\begin{sideways}\textbf{German}\end{sideways}}

 \\ 
& \foreignlanguage{arabic}{\textbf{شكرا}} & \textbf{thanks} & & \foreignlanguage{arabic}{\textbf{ولدان}} & \textbf{two boys}
&&\textbf{Eisenbahnbetrieb}& \textbf{rail operations}
\\
& \foreignlanguage{arabic}{وشكرا} & and thanks && \foreignlanguage{arabic}{ابنان} & two sons
&&Fahrbetrieb& driving
\\

& \foreignlanguage{arabic}{تحياتي} & greetings && \foreignlanguage{arabic}{ولدين} & two boys
&&Reisezugverkehr& passenger trains
\\
& \foreignlanguage{arabic}{شكراً} & thanks + diacritic && \foreignlanguage{arabic}{طفلان} & two children
&&Fährverkehr& ferries
\\
& \foreignlanguage{arabic}{وشكراً} & and thanks + diacritic && \foreignlanguage{arabic}{ابنين} & two sons
&&Handelsverkehr& Trade
\\
& \foreignlanguage{arabic}{مرحبا} & hello && \foreignlanguage{arabic}{ابنتان}  & two daughters
&&Schülerverkehr& students Transport

\\
\cline{2-3}\cline{5-6}\cline{8-9}
\multirow{8}{*}{\begin{sideways}\textbf{Russian}\end{sideways}}& \multicolumn{2}{c}{}   & \multirow{8}{*}{\begin{sideways}\textbf{Chinese}\end{sideways}}&  \textbf{Transliteration}& \multicolumn{1}{c}{} 
&\multirow{8}{*}{\begin{sideways}\textbf{Italian}\end{sideways}} \\
 & \foreignlanguage{russian}{\textbf{Путин}} & \textbf{Putin} && \textbf{dongzhi}  & \textbf{Winter Solstice} &&\textbf{papa}&\textbf{Pope}\\
& \foreignlanguage{russian}{Янукович} & Yanukovych && chunfen & Vernal Equinox &&Papa&Pope\\
& \foreignlanguage{russian}{Троцкий} & Trotsky && xiazhi & Summer solstice &&pontefice&pontiff\\
& \foreignlanguage{russian}{Гитлер} & Hitler  && qiufen & Autumnal Equinox &&basileus&basileus\\
& \foreignlanguage{russian}{Сталин} & Stalin && ziye & Midnight&&canridnale&cardinal\\
& \foreignlanguage{russian}{Медведев} & Medvedev && chuxi & New Year's Eve&&frate&friar\\
\cline{2-3}\cline{5-6}\cline{8-9}

\cline{2-3}\cline{5-6}
\end{tabular}
\end{center}
\end{scriptsize}
\caption{Examples of the nearest five neighbors of every word in several languages. Translation is retrieved from \url{http://translate.google.com}.}
\label{semantic}
\end{table*}

One disadvantage of the approach used by \cite{Collobert:2011:NLP} is that there is no clear stopping criteria for the model training process.
We have noticed that after a few weeks of training, the model's performance reaches the point where there is no significant decrease in the average loss over the development set, and when this occurs we manually stop the training.
An interesting property of this model is that we did not notice any sign of overfitting for large Wikipedias. This could be explained by the infinite amount of examples we can generate by randomly choosing the replacement word in the corrupted phrase.
Figure \ref{error} shows a typical learning curve of the training.
As the number of examples have been seen so far increased both the training error and the development error go down.

\section{Qualitative Analysis}
\label{analysis}
In order to understand how the embeddings space is organized, we examine the subtle information captured by the embeddings through investigating the proximity of word groups.
This information has the potential to help researchers develop applications that use such semantic and syntactic information.
The embeddings not only capture syntactic features, as we will demonstrate in Section \ref{pos}, but also demonstrate the ability to capture interesting semantic information.
Table \ref{semantic} shows different words in several languages.
For each word on top of each list, we rank the vocabulary according to their Euclidean distance from that word and show the closest five neighboring words.

\begin{compactitem}

\item \textbf{French \& Spanish} - Expected groupings of colors and professions is clearly observed.

\item \textbf{English} - The example shows how the embedding space is aware of the name change that happened to a group of Indian cities. ``\emph{Mumbai}" used to be called ``\emph{Bombay}", ``\emph{Chennai}" used to be called ``\emph{Madras} and ``\emph{Kolkata}" used to be called ``\emph{Calcutta}".
On the other hand, ``\emph{Hyderabad}" stayed at a similar distance from both names as they point to the same conceptual meaning.

\item \textbf{Arabic} - The first example shows the word ``\emph{Thanks}". Despite not removing the diacritics from the text, the model learned that the two surface forms of the word mean similar things and, therefore, grouped them together.\
In Arabic, conjunction words do not get separated from the following word.
Usually, "\emph{and thanks}" serves as a letter signature as ``\emph{sincerely}" is used in English.
The model learned that both words \{``\emph{and thanks}", ``\emph{thanks}" \} are similar, regardless their different forms.
The second example illustrates a specific syntactic morphological feature of Arabic, where enumeration of couples has its own form.

\item \textbf{German} - The example demonstrates that the compositional semantics of multi-unit words are still preserved.

\item \textbf{Russian} - The model learned to group Russian/Soviet leaders and other figures related to the Soviet history together.

\item \textbf{Chinese} - The list contains three solar terms that are part of the traditional East Asian lunisolar calendars.
The remaining two terms correspond to traditional holidays that occur at the same dates of these solar terms.

\item \textbf{Italian} - The model learned that the lower and upper cases of the word has similar meaning. 

\end{compactitem}

\section{Sequence Tagging}
\label{results}
Here we analyze the quality of the models we have generated.
To test the quantitative performance of the embeddings, we use them as the sole features for a well studied NLP task, part of speech tagging.

\begin{table*}[ht]
\begin{center}
\begin{tabular}{|ll|rrr|r|}
\hline
\multicolumn{1}{|l}{\multirow{2}{*}{\textbf{Language}}} &
\multicolumn{1}{|l|}{\multirow{2}{*}{\textbf{Source}}} &
\multicolumn{3}{c}{\textbf{Test}} &
\multicolumn{1}{|c|}{\multirow{2}{*}{\textbf{TnT}}}
\\\cline{3-5}

&\multicolumn{1}{|l|}{}& \multicolumn{1}{c}{\textbf{Unknown}} & \multicolumn{1}{c}{\textbf{Known}} & \multicolumn{1}{c|}{\textbf{All}} & \\ \hline
German &Tiger\textsuperscript{$\dagger$} \cite{de}

& 89.17\% & \textbf{98.60}\% & 97.85\% & 98.10\% \\ \hline
Bulgarian & BTB\textsuperscript{$\dagger$} \cite{bg}
& 75.74\% & 98.33\% & 96.33\%  & 97.50\% \\ \hline
Czech & PDT 2.5 \cite{cs}
& 71.98\% & \textbf{99.15}\% &  97.13\%  & 99.10\% \\ \hline
Danish &DDT\textsuperscript{$\dagger$} \cite{da}
& 73.03\% & 98.07\% &  \textbf{96.45\%}  & 96.40\% \\ \hline
Dutch &Alpino\textsuperscript{$\dagger$} \cite{nl}
& 73.47\% & 95.85\% & 93.86\%  & 95.00\% \\ \hline
English  &PennTreebank \cite{en}
& 75.97\% & 97.74\% & \textbf{97.18\%}  &  96.80\% \\ \hline 
Portuguese &Sint(c)tica\textsuperscript{$\dagger$} \cite{pt}
& 75.36\% & 97.71\% & 95.95\%  & 96.80\% \\ \hline
Slovene& SDT\textsuperscript{$\dagger$} \cite{sl}
& 68.82\% & 95.17\% & 93.46\% & 94.60\% \\ \hline
Swedish &Talbanken05\textsuperscript{$\dagger$} \cite{sv}
& 83.54\% & 95.77\% & \textbf{94.68}\%  & 94.70\% \\ \hline
\end{tabular}
\end{center}

\caption{Results of our model against several PoS datasets. The performance is measured using accuracy over the test datasets. Third column represents the total accuracy of the tagger the former two columns reports the accuracy over known words and OOV words (unknown). The results are compared to the TnT tagger results reported by \cite{universal}.\\
\footnotesize{\textsuperscript{$\dagger$}CoNLL 2006 dataset}}
\label{pos}
\end{table*}

To demonstrate the capability of the learned distributed representations in extracting useful word features, we train a PoS tagger over the subset of languages that we were able to acquire free annotated resources for.
We choose our tagger for this task to be a neural network because it has a fast convergence rate based on our initial experiments.

The part of speech tagger has similar architecture to the one used for training the embeddings.  However we have changed some of the network parameters, specifically, we use a hidden layer of size $300$ and learning rate of $0.3$.
The network is trained by minimizing the negative of the log likelihood of the labeled data.
To tag a specific word $w_i$ we consider a window with size $2n$ where $n$ in our experiment is equal to $2$.
Equation \ref{cat} shows how we construct a feature vector $F$ by concatenating ($\oplus$) the embeddings of the words occurred in the window, where $C$ is the matrix that contains the embeddings of the language vocabulary.
\begin{equation}
F = \bigoplus_{j=i-2}^{i+2}{C[w_j]}
\label{cat}
\end{equation}
The feature vector will be fed to the network and the error will back propagated back to the embeddings.

The results of this experiment are presented in Table \ref{pos}. We train and test our models on the universal tagset proposed by \cite{universal}.
This universal tagset maps each original tag in a treebank to one out of twelve general PoS tags. This simplifies the comparison of classifiers performance across languages.
We compare our results to a similar experiment conducted in their work, where they trained a TnT tagger \cite{tnt} on several treebanks.
The TnT tagger is based on Markov models and depends on trigram counts observed in the labeled data. It was chosen for its fast speed and (near to) state-of-the-art accuracy, without language specific tuning.

The performance of embeddings is competitive in general. Surprisingly, it is doing better than the TnT tagger in English and Danish.
Moreover, our performance is so close in the case of Swedish.
This task is hard for our tagger for two reasons.
The first is that we do not add OOV words seen during training of the tagger to our vocabulary.
The second is that all OOV words are substituted with one representation, $\langle$UNK$\rangle$ and there is no character level information used to inform the tagger about the characteristic of the OOV words.

On the other hand, the performance on the known words is strong and consistent showing the value of the features learned about these words from the unsupervised stage.
Although the word coverage of German and Czech are low in the original Wikipedia corpora (See Table \ref{table:langs}), the features learned are achieving great accuracy on the known words.
They both achieve above 98.5\% accuracy. 
It is noticeable that the Slovene model performs the worst, under both known and unknown words categories.
It achieves only 93.46\% accuracy on the test dataset.
Given that the Slovene embeddings were trained on the least amount of data among all other embeddings we test here, we expect the quality to go lower for the other smaller Wikipedias not tested here.

In Table \ref{table:poscoverage}, we present how well the vocabulary of each language's embeddings covered the part of speech datasets.
The datasets come from a different domain than Wikipedia, and this is reflected in the results.

In Table \ref{table:pretraining}, we present the results of training the same neural network part of speech tagger without using our embeddings as initializations.  We found that the embeddings benefited all the languages we considered, and observed the greatest benefit in languages which had a small number of training examples.  We believe that these results illustrate the performance 

\begin{table}[ht]
\centering
\begin{tabular}{|l|c|c|}
\hline
\multirow{2}{*}{\textbf{Language}} & \multicolumn{1}{c|}{\textbf{\% Token}} & \multicolumn{1}{c|}{\textbf{\% Word }} \\
 & \multicolumn{1}{c|}{\textbf{Coverage}} & \multicolumn{1}{c|}{\textbf{Coverage}} \\ \hline
Bulgarian&	94.58&	77.70\\
Czech&	95.37&	65.61\\
Danish&	95.41&	80.03\\
German&	94.04&	60.68\\
English&	98.06&	79.73\\
Dutch&	96.25&	77.76\\
Portuguese&	94.09&	72.66\\
Slovene&	95.33&	83.67\\
Swedish&	95.87&	73.92\\ \hline
\end{tabular}

\caption{Coverage statistics of the embedding's vocabulary on the part of speech datasets after normalization.  Token coverage is the raw percentage of words which were known, while the Word coverage ignores repeated words.}
\label{table:poscoverage}
\end{table}

\begin{table}[ht]
\centering
\begin{tabular}{|l|c|c|}
\hline
\multirow{2}{*}{\textbf{Language}} & \multicolumn{1}{c|}{\textbf{\# Training }} & \multicolumn{1}{c|}{\textbf{Accuracy}} \\
 & \multicolumn{1}{c|}{\textbf{Examples}} & \multicolumn{1}{c|}{\textbf{Drop}}  \\ \hline
Bulgarian&	200,049&	-2.01\%\\
Czech&	1,239,687&	-0.86\%\\
Danish&	96,581&	-1.77\%\\
German&	735,826&	-0.89\%\\
English&	950,561&	-0.25\%\\
Dutch&	208,418&	-1.37\%\\
Portuguese&	212,749&	-0.91\%\\
Slovene&	27,284&	-2.68\%\\
Swedish&	199,509&	-0.82\%\\ \hline
\end{tabular}

\caption{Accuracy of randomly initialized tagger compared to our results. Using the embeddings was generally helpful, especially in languages where we did not have many training examples.
The scores presented are the best we found for each language (languages with more resources could afford to train longer before overfitting). }
\label{table:pretraining}
\end{table}

\section{Conclusion}
\label{conc}
Distributed word representations represent a valuable resource for any language, but particularly for resource-scarce languages.
We have demonstrated how word embeddings can be used as off-the-shelf solution to reach near to state-of-art performance over a fundamental NLP task, and we believe that our embeddings will help researchers to develop tools in languages with which they have no expertise.

Moreover, we showed several examples of interesting semantic relations expressed  in the embeddings space that we believe will lead to interesting applications and improve tasks as semantic compositionality.

While we have only considered the properties of word embeddings as features in this work, it has been shown that using word embeddings in conjunction with traditional NLP features can significantly improve results on NLP tasks \cite{turian2010word,Collobert:2011:NLP}.  With this in mind, we believe that the entire research community can benefit from our release of word embeddings for over 100 languages.

We hope that these resources will advance the study of possible pair-wise mappings between embeddings of several languages and their relations.
Our future work in this area includes improving the models by increasing the size of the context window and their domain adaptivity through incorporating other sources of data.
We will be investigating better strategies for modeling OOV words.  We see improvements to OOV word handling as essential to ensure robust performance of the embeddings on real-world tasks.

\section*{Acknowledgments}
This research was partially supported by NSF Grants DBI-1060572 and IIS-1017181, with additional support from TexelTek.

\bibliographystyle{acl}
\bibliography{refs}
\end{document}